%% file: nips20.tex
\documentclass{article}
\PassOptionsToPackage{numbers, compress}{natbib}

\usepackage[final]{neurips_2020}

\usepackage[utf8]{inputenc} 
\usepackage[T1]{fontenc}    
\usepackage{hyperref}       
\usepackage{url}            
\usepackage{booktabs}       
\usepackage{amsmath}
\usepackage{algorithm}
\usepackage{algcompatible}
\usepackage{draftfigure}
\usepackage{subcaption}
\usepackage{xcolor}
\usepackage{amsfonts}       
\usepackage{nicefrac}       
\usepackage{microtype}      
\usepackage{natbib}
\usepackage{enumitem}
\DeclareMathOperator*{\argmax}{\arg\max}
\newcommand{\cut}[1]{}

\newcommand{\doublecheck}[1]{\textcolor{blue}{#1}}
\newcommand{\todo}[1]{\textcolor{red}{(#1)}}
\newcommand{\keypoint}[1]{\textbf{#1}\quad}
\newcommand{\nameS}{P2PDRL}
\newcommand{\modelS}{P2PDRL}

\title{Robust Domain Randomised Reinforcement Learning through Peer-to-Peer Distillation}
\author{
    Chenyang Zhao\\
    School of Informatics\\
    University of Edinburgh\\
    \texttt{c.zhao@ed.ac.uk} \\
    \And
    Timothy M. Hospedales\\
    School of Informatics\\
    University of Edinburgh\\
    \texttt{t.hospedales@ed.ac.uk} \\
    
}

\begin{document}

\maketitle

\begin{abstract}
    In reinforcement learning, domain randomisation is an increasingly popular technique for learning more general policies that are robust to domain-shifts at deployment. However, naively aggregating information from randomised domains may lead to high variance in gradient estimation and unstable learning process. To address this issue, we present a peer-to-peer online distillation strategy for RL termed \nameS{}, where multiple workers are each assigned to a different environment, and exchange knowledge through mutual regularisation based on Kullback–Leibler divergence. Our experiments on continuous control tasks show that  \nameS{} enables robust learning across a wider randomisation distribution than baselines, and more robust generalisation to new environments at testing. 
\end{abstract} 

\section{Introduction}

Deep reinforcement learning (RL) has been successfully applied to various tasks, including Go \cite{silver2016mastering}, Atari \cite{mnih2015dqn} and robot control tasks \cite{schulman2017proximal}, etc. However, a growing body of research has shown that it remains challenging to learn deep RL agents that are able to generalise to new environments \cite{cobbe2018quantifying}. Agents can `overfit' to the training environments visual appearance \cite{tobin2017domain,cobbe2018quantifying} physical dynamics \cite{packer2018assessing,zhao2019investigating} or even specific training seeds \cite{zhang2018dissection}. If there is a new environment, or domain-shift, between training and testing, RL tend to under-perform significantly. This problem has longer been appreciated, and is particularly salient, in the field of robotics, where there is an inevitable \textit{reality gap}  \cite{tobin2017domain,koos2012transferability} between the simulated environments used for training and deployment in the real-world. These issues have motivated an important line of research in improving zero-shot domain transfer \cite{cobbe2018quantifying,andrychowicz2020learning}, i.e., to  learn generalised policies from training domains that can be directly and successfully deployed in unseen testing domains without further learning or adaptation. 

One common strategy to improve policy generalisation is to apply \textit{domain randomisation} during training. With domain randomisation, we first generate a random set of, or distribution over, domains with different properties such as observation function \citep{tobin2017domain}, dynamics \citep{peng2018sim}, or both \citep{andrychowicz2020learning}. Then a policy is trained with this distribution, for example by drawing a new random sample from the domain distribution during each learning episode. If this distribution of training domains is rich enough to include within its support properties of the real-world or target environment -- and if a policy can be successfully trained to fit the entire training domain distribution -- then that policy can be successfully deployed in the target environment without adaptation. A challenge with  domain randomisation is that  training a single model on a wide distribution of domains usually leads to high variance gradient estimates, making the RL policy optimisation problem challenging \citep{mehta2019active}. To ameliorate this issue, several studies have proposed distillation techniques where an ensemble of workers are each trained locally in a different domain -- where gradients are lower variance, and hence RL is more stable. The local workers are then distilled into a single global policy that should perform across all domains \citep{ghosh2017divide, teh2017distral, rusu2015policy, parisotto2015actor, czarnecki2019distilling}. Distilling local policies into a global policy \cut{\doublecheck{(periodically, or once)}}with a supervised objective provides a more stable way to learn from a broad distribution over domains and ultimately enable better generalisation to held-out testing domains --  albeit at some additional computational cost \citep{ghosh2017divide}. 

This work continues this line of investigation. Our main contribution is to show that this centralised distillation is  unnecessarily, and indeed sub-optimal. In particular, we propose an online distillation framework, where each worker both learns to optimise performance in a local domain and also mimics its peers from other domains with peer-to-peer distillation. More specifically, inspired by deep mutual learning \citep{zhang2018deep, anil2018large}, we train each worker with two losses: a conventional RL loss and a distillation loss that measures the similarity in predictions between the local worker and the others. In this work, we use expected Kullback–Leibler (KL) divergence between predicted distributions as the objective to optimise for information exchange. This online distillation framework is named \textit{Peer-to-Peer Distillation Reinforcement Learning} (\nameS{}). We empirically show that, compared to conventional baselines, and offline distillation alternatives, e.g. divide-and-conquer (DnC) reinforcement learning \citep{ghosh2017divide}, \nameS{} achieves more competitive learning performance without the additional cost of a centralised distillation step. Overall \nameS{} learns more quickly, succeeds to learn on a wider distribution of domains, and transfers better to novel testing environments compared to competitors. 

The key contributions of this work are summarised as:
\begin{itemize}
    \item We propose \nameS{}, an online distillation framework for deep RL agents that uses an ensemble of local workers that learn with local experience only, while sharing knowledge via regularisation on the statistical distance of their prediction distribution;
    \item Our experiments show that \nameS{} can achieve more effective learning on a wider distribution of training environments compared to competitors, without distilling to a centralised policy. All workers show increased robustness to the domain shift of novel environments. 
\end{itemize}

\section{Related Work}
\keypoint{Domain Generalisation in Deep RL}
A growing body of work has discussed generalisation problems in reinforcement learning. In the context of tasks with discrete and continuous observation/action spaces, \cite{cobbe2018quantifying} and \cite{zhang2018dissection} concluded that deep RL policies could easily overfit to random seeds used during training and underperform during testing. In the context of domain-shift between training and testing, \cite{zhao2019investigating} and \cite{packer2018assessing} further highlighted the problem that deep RL agents often fail to generalise, e.g., across different friction coefficients or wind conditions.

\begin{figure}[t]
    \centering
    \begin{subfigure}{.30\textwidth}
    \centering
    \includegraphics[width=\textwidth]{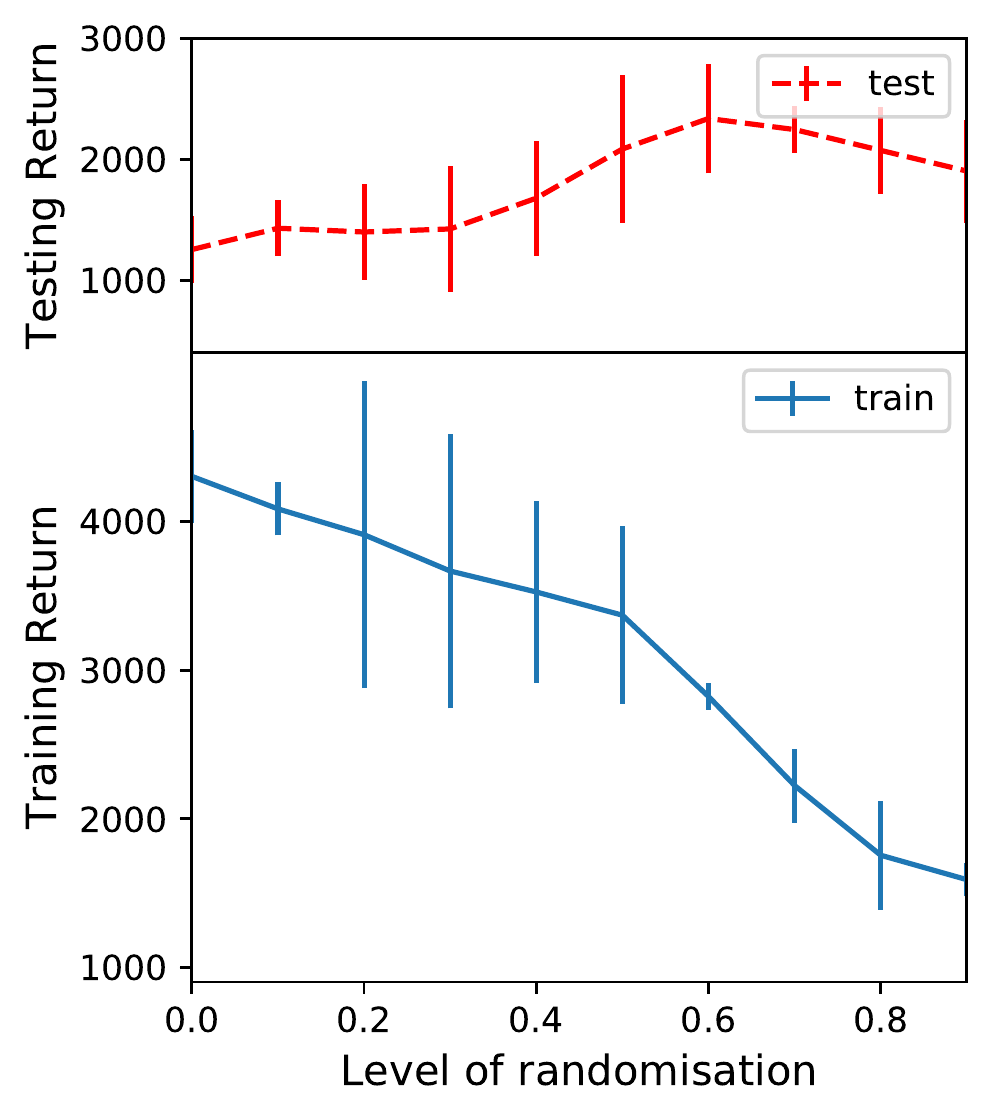}
    \caption{Walker task}
    \end{subfigure}
    \hspace{1.0cm}
    \begin{subfigure}{.30\textwidth}
    \centering
    \includegraphics[width=\textwidth]{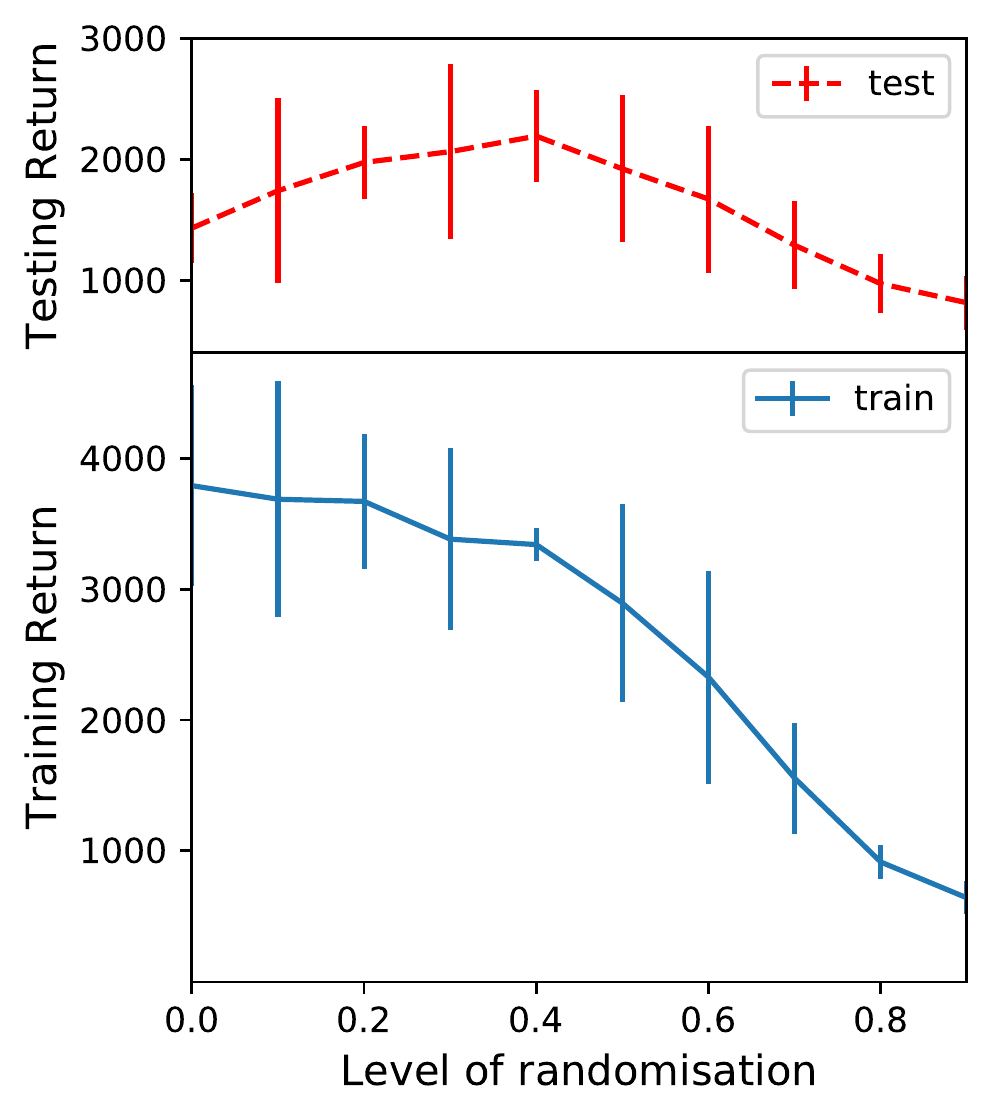}
    \caption{Ant task}
    \end{subfigure}
    \caption{Comparisons between asymptotic training performance and testing performance in a hidden testing domain, both as a function of the diversity of training distributions. Policies are trained with PPO. Results are averaged over 8 different random seeds.}
    \label{fig: unstable learning}
\end{figure}

To improve RL agents' performance in testing environments, several \textit{adaptation} methods have been proposed \cite{gupta2017learning, rusu2016sim}. These use data from testing domains and aim to perform sample efficient adaptation. Recent advances \cite{clavera2019learning,ritter2018episodicMeta} perform meta-learning on training environments to learn how to adapt efficiently to testing environments. 
Where direct deployment to target domains without adaptation is desired or necessary, a common approach is \textit{domain randomisation}. Here the training simulator is setup to provide a diverse array of environments for learning in terms of observations \citep{sadeghi2016cad2rl, tobin2017domain, andrychowicz2020learning} or dynamic perturbations \citep{tan2018sim}. A key challenge here is that, in absence of a known and precisely-specified model of the testing domain, one must define a rather wide distribution of training domains to be confident that its span includes likely testing domains. However, learning from such a diverse training signal is extremely challenging in RL, and often leads to unstable learning and poor convergence \cite{mehta2019active,Yu2019}, as we also illustrate here in Fig.~\ref{fig: unstable learning}. 

In this paper we address this dichotomy between overfitting of agents to individual domains, and inability of agents to fit to a wide distribution of domains. Our framework enables state of the art on-policy methods \cite{schulman2017proximal} to effectively and stably learn a wide distribution of domains by performing local RL and global knowledge exchange via distillation. 



\keypoint{Policy Distillation}
The notion of model distillation \cite{hinton2014distilling} has increasingly been applied to policy distillation in reinforcement learning. It is often used for learning from multiple source tasks, where a student agent is trained to mimic the behaviour of one or multiple teacher policies \citep{czarnecki2019distilling}. Several early studies \cite{rusu2015policy,parisotto2015actor} assumed pre-trained teacher policies and trained one student to match the state-dependent probability of the teacher's predicted actions. Later methods  \cite{teh2017distral} learned a global policy in parallel with learning multiple local policies: the global policy is trained to distil from local individuals while local policies learning is regularised by the global policy. Finally, \cite{ghosh2017divide} proposed to alternate between training local agents regularised by a global policy,  distilling a central global policy, and periodically resetting local agents to the distilled global policy. 

Compared to these methods, we propose a online peer-to-peer distillation framework that avoids explicitly distilling to a global policy. Instead, we learn a cohort of workers collaboratively by performing local RL and global peer-to-peer knowledge distillation. {Peers use their own local states for distillation, thus significantly simplifying data exchange between workers compared to alternatives such as \cite{ghosh2017divide}.}
This idea of online distillation has also been explored in supervised learning to achieve learning better performance in single-task learning \citep{zhang2018deep} or to scale single-task supervised learning to large datasets through distributed computing  \citep{anil2018large}. To our knowledge, it has not been explored in RL, or as a tool to enable learning on a wider distribution of training domains, and hence improved robustness and cross-domain generalisation. 

\section{Methodology}
 
\subsection{Preliminaries}
An episodic Markov decision process (MDP) is defined by $\langle \mathcal{S}, \mathcal{A}, R, T, \gamma \rangle$, where $\mathcal{S}$ is the state space, $\mathcal{A}$ is the action space, $R: \mathcal{S} \times \mathcal{A} \rightarrow \mathbb{R}$ is the reward function, $T: \mathcal{S} \times \mathcal{A} \rightarrow \mathcal{S}$ is the transition function, and $\gamma$ is the discount factor. The objective of a learning agent is to learn a stochastic policy $\pi^*: \mathcal{S}\times\mathcal{A}\rightarrow\mathbb{R}$ that maximises the expected cumulative return:~ $\pi^*~=~\argmax_{\pi}\mathbb{E}_{\pi}\sum_{t=0}^\infty \gamma^t R(s_t,a_t)$. 

We denote the parameters that describe a domain as $\xi$. In domain randomisation, each training domain is randomly sampled with domain parameters $\xi$ from a pre-defined set $\Xi$. 
Thus, the modified objective function becomes 
\begin{equation}
    \pi_* = \argmax_\pi \mathbb{E}_{\xi\sim\Xi}\mathbb{E}_\pi \sum_{t=0}^\infty \gamma^t R(s_{t,\xi},a_{t,\xi}).
    \label{eq: rl loss}
\end{equation}
Policies trained with a diverse set of domains should generalise better to unseen testing domains \cite{andrychowicz2020learning}. However, in practice,  diverse training domains lead to high variance in gradients \citep{mehta2019active}, {thus leading to poor learning behaviour}. In practice, this manifests as a significant drop in training performance as diversity of training domains increases. In such cases, the effect of the policy's inability to fit the training task at all dominates its greater exposure to diverse environments due to domain randomisation. Thus performance in both training and testing domains is poor. 

This challenge is illustrated in Fig.~\ref{fig: unstable learning}, which shows the performance of an agent during training as well as its performance when tested on a held-out domain -- all plotted  as a function of training set diversity. As the diversity of training domains increases from zero, we see that training performance decreases continually as it becomes harder for a single policy to fit an increasingly diverse distribution of environments (Fig.~\ref{fig: unstable learning}, below). More interestingly, we also observe that testing performance initially increases, as the increased domain randomisation benefits robustness to the held out domain (Fig.~\ref{fig: unstable learning}, above). However as training diversity continues to increase, testing performance starts to drop, as the stronger domain randomisation challenge makes it difficult for the policy to learn the task at all. 


\subsection{Online Distillation}
To avoid training with data sampled from all domains, we propose to assign each worker to a domain, and train it with data from its local domain only. Alongside the conventional RL loss, workers are regularised by each other through an online distillation loss, which encourages all workers to act similarly across all domains given the same state input.

\begin{figure}[t]

    \centering
    \includegraphics[width=0.80\textwidth]{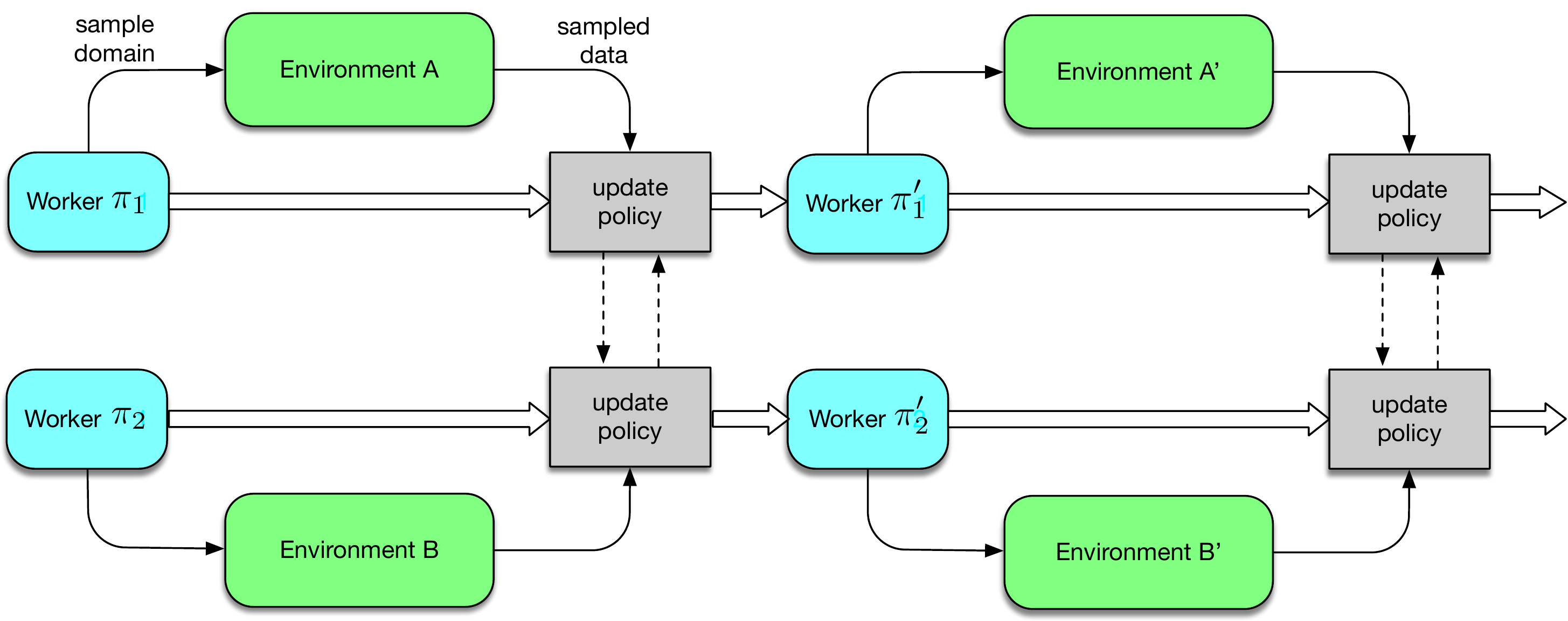}
    \caption{Schematic of \nameS{} with two workers. At each iteration, a worker is randomly assigned to a domain, and updates its policy by local RL, while being regularised by its peers learning in other domains. Dotted lines indicate matching prediction distribution by KL divergence based distillation.}
    \label{fig: mutual}
\end{figure}

\keypoint{Local Optimisation}
We formulate the proposed peer-to-peer distillation method with a cohort of $K$ workers, as illustrated in Fig.~\ref{fig: mutual}. At each iteration, each worker randomly samples a domain and generates trajectory data by following its local policy. The objective function includes a conventional RL loss and a KL divergence based distillation loss that matches predicted distributions across all workers. In particular, we use proximal policy optimisation (PPO) \citep{schulman2017proximal}, a state of the art on-policy deep reinforcement learning algorithm, as the base optimiser. A surrogate loss is used in PPO: 
\begin{equation}
    \mathcal{L}_{\text{PPO}}(\theta; \xi) = \mathbb{E}_{s_t,a_t} \bigg[ \min\big(\frac{\pi_\theta(a_t|s_t)}{\pi_{\theta_{\text{old}}(a_t|s_t)}}A_{t,\xi}, \text{clip}(\frac{\pi_\theta(a_t|s_t)}{\pi_{\theta_{\text{old}}(a_t|s_t)}}, 1-\epsilon, 1+\epsilon)A_{t,\xi} \big)\bigg],
\end{equation}
where $\epsilon$ is the hyperparameter, and $A_{t,\xi}$ is the estimated advantage in domain $\xi$ at timestep $t$. The critic networks, parameterised by $\phi_i$, are trained locally with conventional MSE loss, where target values are computed through Bellman equations using local trajectory data. 


\keypoint{Peer-to-Peer Distillation}
For online distillation across different workers, we minimise KL divergence between workers' predictions. More specifically, for the $i$-th worker, 
\begin{equation}
    \mathcal{L}^i_{\text{dis}}(\theta) = \frac{1}{K-1}\sum_{k}^{k\neq i} \mathbb{E}_{\pi_i, \xi_i} \big[ D_{\text{KL}}(\pi_{\theta_i}(\cdot|s) || \pi_{\theta_k}(\cdot|s)) \big],
\end{equation}
where $\mathbb{E}_{\pi_i, \xi_i}(\cdot)$ is the expectation over trajectory data generated in domain $\xi_i$ with policy $\pi_i$. 

\keypoint{Discussion} Our overall  algorithm is summarised in Fig.~\ref{fig: mutual} and  Alg.~\ref{alg: ppo}.  Note that information is only exchanged in distillation steps between workers. Different from DnC \citep{ghosh2017divide}, we do not require local workers to have access to trajectory data sampled by peers; only model parameters need to be shared between workers. This simplifies data exchange and enables more efficient implementation. Furthermore, we do not distill local policies to a central global policy. Although each worker optimises locally, over time they \emph{all} become domain invariant due to peer-to-peer knowledge exchange.

\cut{\keypoint{Extension to Asynchronous Learning} 
{In the vanilla version of our method, workers synchronise between data collection steps, and access peer policies directly to perform peer-to-peer distillation-based knowledge exchange.} 
Recent work suggest that distillation-based knowledge exchange is robust to the use of out-of-date copies of individual workers during online distillation \citep{anil2018large}. Similarly in the context of RL, we can take asynchronous update steps, i.e., having local workers train in parallel and periodically checkpoint their parameters to peers. 
The timeline of synchronous versus asynchronous updates is shown in Fig.~\ref{fig: sync timeline}-\ref{fig: async timeline}. 
As illustrated in these figures, such asynchronous updates help to reduce training time by reducing data exchange between local workers, and eliminating waiting time for synchronisation. Such asynchronous updates are also more amenable to distributed implementation with local workers executing on different nodes.}

\begin{algorithm}[t]
\caption{Online Peer-to-Peer Distillation Reinforcement Learning with PPO}
\label{alg: ppo}
\begin{algorithmic}[1]
\STATE \textbf{Input:} Distribution over domains $\Xi$, number of workers $K$, hyperparameter $\alpha$, max timesteps $T$.
\STATE \textbf{Initialise:} Initial actor $\theta_{0}$, initial critic $\phi_{0}$.
\STATE \textbf{Initialise:} every worker with $\theta_0, \phi_0$
\WHILE{not converge}
\FOR{$i=1,2,...,K$}
\STATE Sample a domain $\xi_i\sim\Xi$ for worker $i$
\STATE Collect trajectory data $\tau_i$ following policy $\pi_i$ for $T$ timesteps in domain $\xi_i$.
\ENDFOR
\FOR{$1,2,..., $ max epoch number}
\FOR{$1,2,..., $ num of minibatches per epoch}
\FOR{$i = 1$ \textbf{to} $K$}
\STATE Sample minibatch $\mathcal{B}_i$ from $\tau_i$
\STATE $\theta_i \leftarrow \theta_i - \nabla_{\theta_i} [\mathcal{L}_{\text{PPO}}(\mathcal{B}_i,\theta_i) + \alpha \mathcal{L}^i_{\text{dis}}(\mathcal{B}_i,\theta_i)]$
\STATE Compute target value $V^{targ}$ with Bellman equation.
\STATE Update $\phi_i$ with MSE loss $(V_n - V^{\operatorname{targ}})^2$.
\ENDFOR
\ENDFOR
\ENDFOR
\ENDWHILE
\end{algorithmic}
\end{algorithm}
\section{Experiments and Results}
\subsection{Experiment Setup}
We focus our analysis on five continuous control tasks from OpenAI Gym, including \textit{Walker, Ant, Humanoid, Hopper, HalfCheetah} \citep{1606.01540}. All tasks are simulated in MuJoCo. Illustrative figures of tasks are included in Fig.~\ref{fig: table fig}. To generate different domains, we change wind condition, gravity constant, friction coefficient, robot mass and initial state distribution in simulation. We summarise the diversity of randomised distribution as a scalar $\epsilon\in[0,1]$. Higher $\epsilon$ means more diverse distributions. Given a specific $\epsilon$, dynamic parameters are randomly sampled from a uniform distribution defined by $\epsilon$. For example, robot mass in \textit{Walker} task $m$ is sampled from $m \sim \mathcal{U}(m_0(1-0.5\epsilon), m_0(1+0.5\epsilon))$. Detailed task descriptions are listed in the Appendix.

Our experiments are designed to address the following questions:

\textbf{Q1:} How does \nameS{} compare to baselines in sample efficiency and asymptotic return?

\textbf{Q2:} Does  \nameS{} learn policies that are more robust against domain shifts than competing  methods? 

We compare our \nameS{}, with the following prior methods as baselines:

\textbf{PPO} \citep{schulman2017proximal}. PPO provides the state of the art on-policy Deep RL method, which is also used as the basic learning algorithms across all settings. At each iteration, agents are optimised with the union of datasets from all sampled domains.

\textbf{Distributed PPO} \cite{heess2017emergence}. In contrast to PPO, where one agent is optimised by the rich data set, in this setting, workers gather gradient information with local data and a central learner updates a global policy with the average of gradients across all domains.

\textbf{Distral} \cite{teh2017distral}. Originally focused on knowledge transferring when learning multiple tasks, Distral trains an ensemble of task-specific policies, constrained against a single global policy at each gradient step. The global policy is trained with supervised learning, to distil the common behaviours from task-specific policies.

\textbf{DnC} \cite{ghosh2017divide}. Originally focused on providing robustness to choice of initial state in single domain learning, DnC partitions the initial state distribution into multiple sub-domains, and alternates between local policy optimisation steps and global distillation steps. In this work, we go beyond initial states, and train agents in the domains with various sets of dynamics parameters. Moreover, we do not require domain knowledge to partition the distribution. Instead, domains are sampled from the same distribution for all workers. 

\input{sec4/table_figure}

\keypoint{Implementation Details} To demonstrate the performance of our proposed method, we take $K=2$ workers in parallel, regularised by each other. Except where noted otherwise, we control the \emph{total} number of timesteps to compare sample efficiency fairly: At each iteration, the vanilla PPO samples 4096 steps; and each worker in other algorithms samples 2048 steps. In DnC and \nameS{}, half the number of samples are used by each actor to train locally. For the implementation of actor and critic networks, we use two separate MLP networks with two hidden layers, 64 units in each layer. For each task, we run a hyperparameter sweep for each method, and report the best performing agents. All results are averaged over 8 random seeds. Full implementation details are listed in the Appendix.  
\subsection{Training and Generalisation Performance}
\keypoint{Training Efficiency Under Domain Randomisation} 

As training distribution, we take $\epsilon^{tr}=0.2$ for all five tasks. From the learning curves shown in Fig.~\ref{fig: table fig}(middle), we can see that \nameS{} generally outperforms the baseline methods in terms of asymptotic performance across all 5 tasks, and is able to learn more sample-efficiently in some tasks, such as \textit{Ant} and \textit{HalfCheetah}. 

\keypoint{Generalisation to Novel Environments} 
We next evaluate the generalisation performance of the policy learned above, by evaluating it on novel testing domain with  different randomisation strengths $\epsilon^{te}$. Fig.~\ref{fig: table fig}(right) shows that \nameS{} learned policies are able generally to generalise better against domain shifts to novel environments. 

\begin{figure}[t]
     \centering
    \begin{subfigure}{.30\textwidth}
    \includegraphics[width=\textwidth]{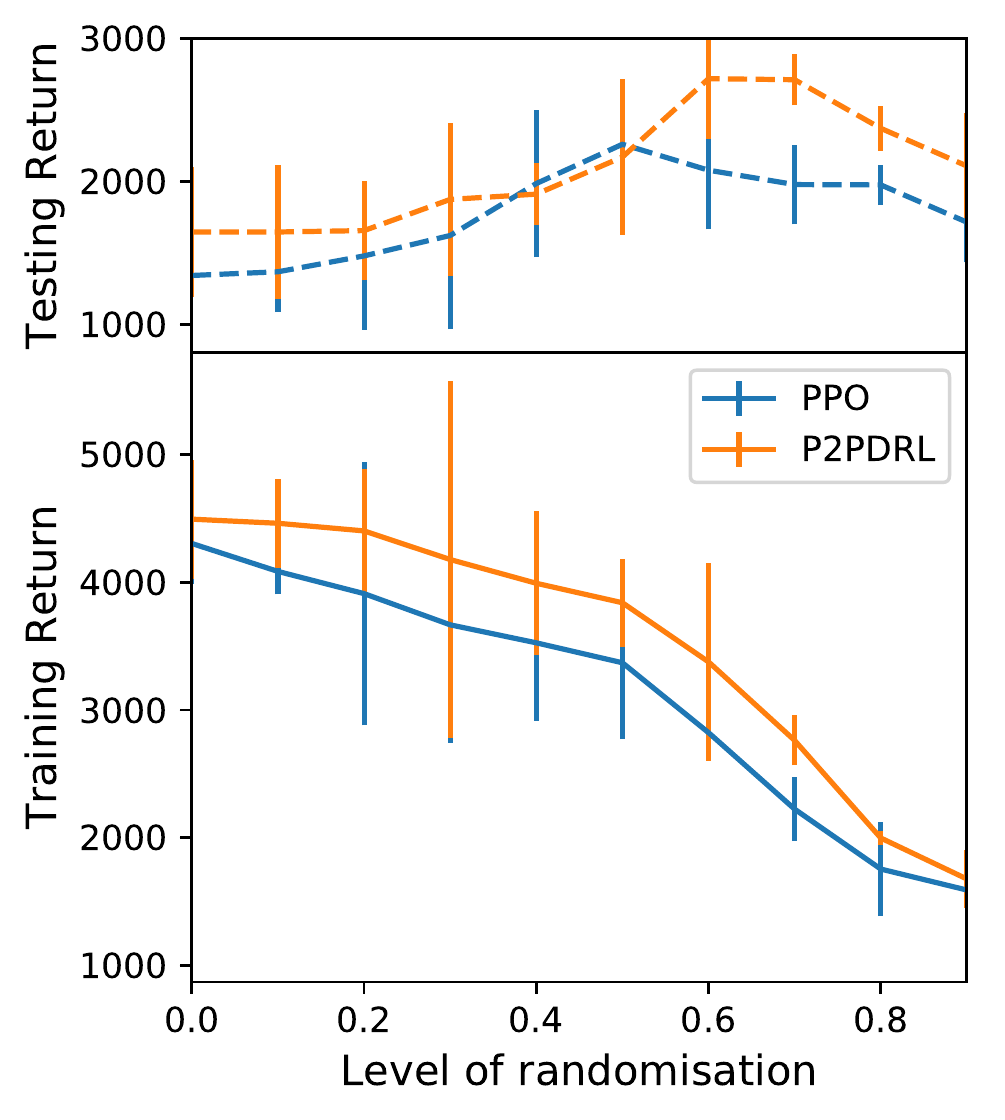}
    \caption{Walker task}
    \end{subfigure}
    \hspace{1cm}
    \begin{subfigure}{.30\textwidth}
    \includegraphics[width=\textwidth]{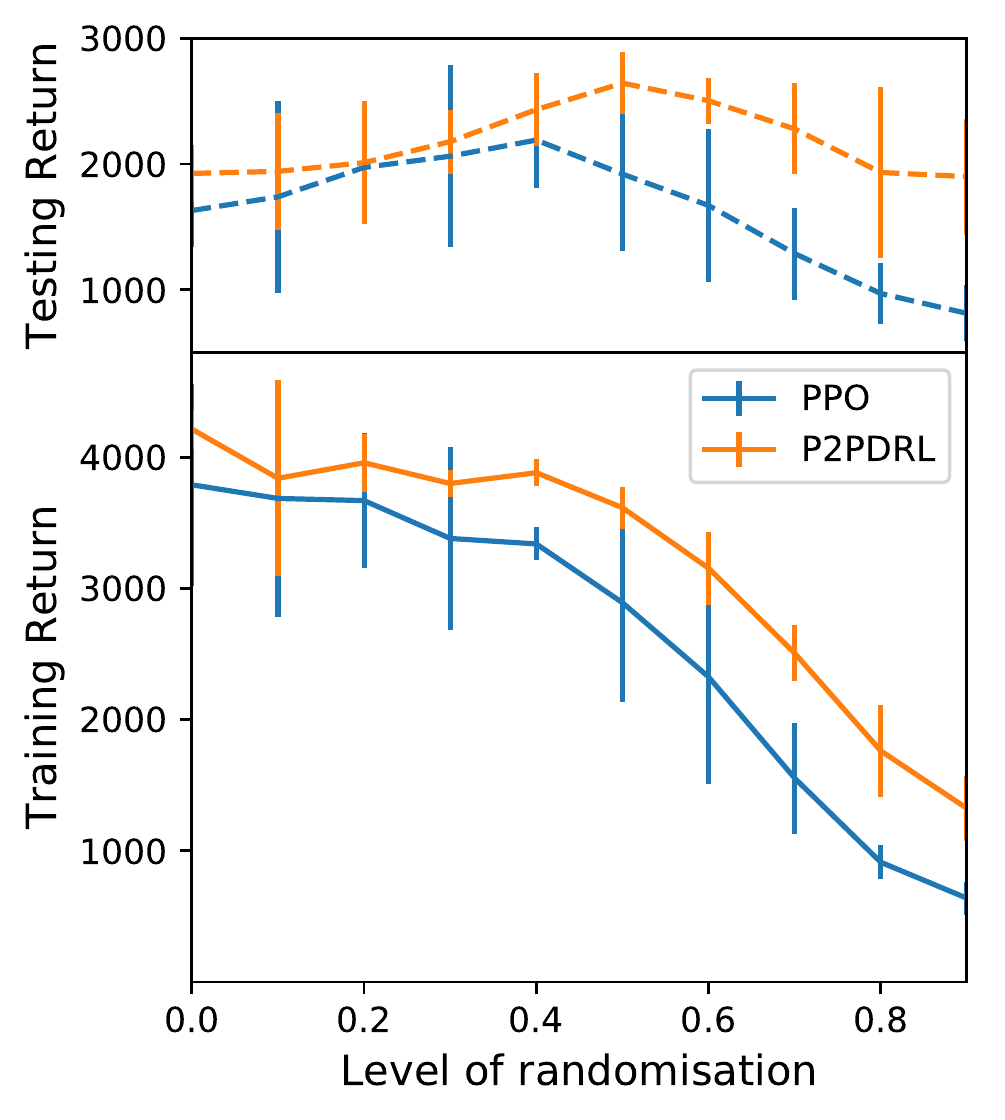}
    \caption{Ant task}
    \end{subfigure}
    \caption{Comparisons between asymptotic training and testing performance, both as a function of the diversity of the training randomisation distribution. Policies are trained with different randomisation strengths $\epsilon^{tr}$ and tested with a pre-defined held-out testing domain.}
    \label{fig: unstable learning distil}
    \vspace{-0.1cm}
\end{figure}

\keypoint{Dependence of Testing Performance on Randomisation Strength in Training} 
We also demonstrate asymptotic training return of a policy and its generalisation performance as a function of range of training domains, as initially motivated in Fig.~\ref{fig: unstable learning}. We train our policy with different levels of randomisation $\epsilon^{tr}$ and test all learned policies in a pre-defined target domain where $\epsilon^{te}=0.5$. As shown in Fig.~\ref{fig: unstable learning distil}, \modelS{}  generally outperforms vanilla PPO, but their performance levels decrease in tandem as the training distribution becomes more diverse and harder to fit with a single model. Crucially, as discussed earlier, the generalisation performance of PPO in terms of testing return drops rapidly after around $\epsilon^{tr}=0.5$. Meanwhile the testing return of \modelS{} is stable up to a much higher level of training domain randomisation. 

\section{Conclusion}
Generalisation of trained agents to novel domains is an crucial but lacking capability in today's RL. We propose an online distillation based reinforcement learning algorithm P2PDRL, consisting of domain-local RL steps and global knowledge exchange through peer-to-peer distillation. We empirically show that P2PDRL provides improved generalisation to novel domains via  stable training on a wide range of randomised domains. 


\cut{\section*{Broader Impact} Existing reinforcement learning methods sometimes disappoint in deployment due to the lack of robustness of their policies after training in simulation. This work provides an improved algorithm for training robust policies. The use of reinforcement learning in robotics is rare today. However, ultimately continued improvements in the ability to train robust policies in simulation could lead to greater use of RL to train robot control skills that are hard to design manually, and hence lead to societal benefits by improving the efficacy of state of the art robotics, and increasing the breadth of tasks that can benefit from automation. }

\bibliography{nips20}

\newpage

\input{supplementary}
\end{document}

%% file: sec4/table_figure.tex
\begin{figure}[p]
\begin{subfigure}{0.32\textwidth}
    \centering
    \includegraphics[width = 0.75\textwidth]{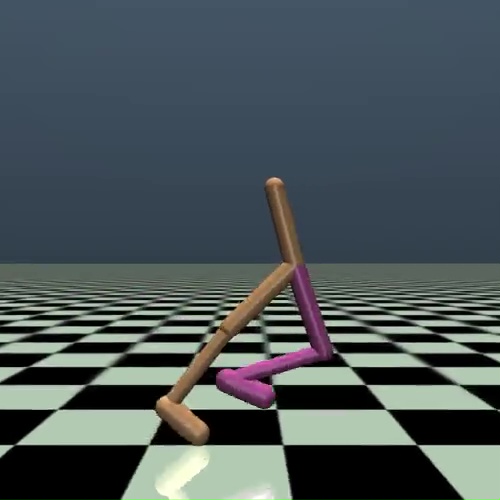}
    \caption{Walker task}
\end{subfigure}
\begin{subfigure}{0.32\textwidth}
    \centering
    \includegraphics[width = \textwidth]{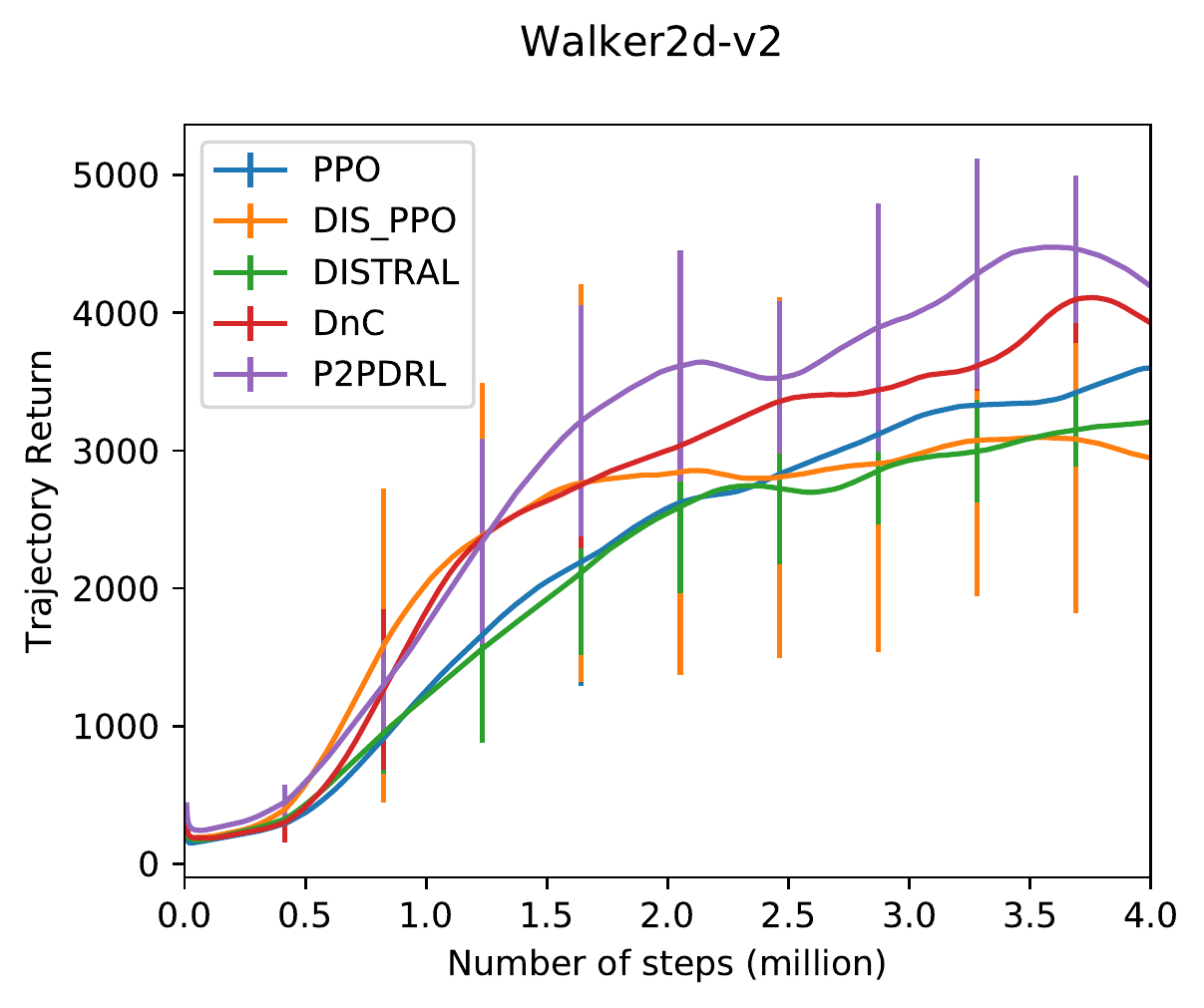}
    \caption{Training return on Walker}
\end{subfigure}
\begin{subfigure}{0.32\textwidth}
    \centering
    \includegraphics[width = \textwidth]{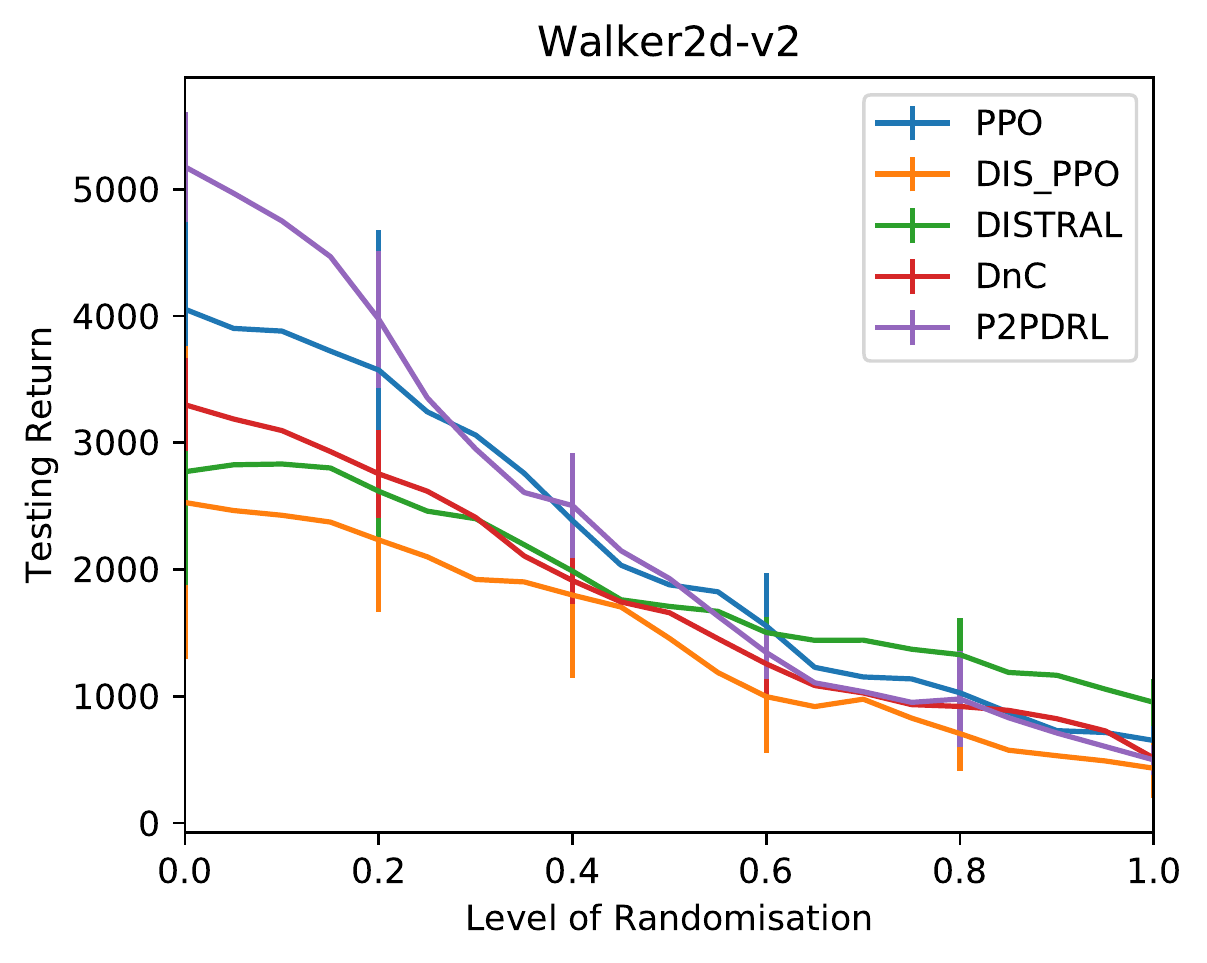}
    \caption{Testing return on Walker}
\end{subfigure}

\begin{subfigure}{0.32\textwidth}
    \centering
    \includegraphics[width = 0.75\textwidth]{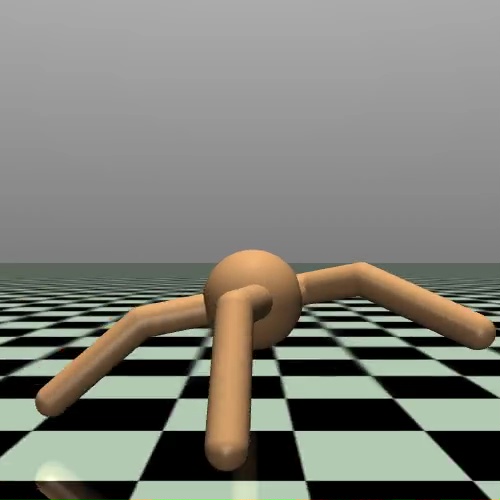}
    \caption{Ant task}
\end{subfigure}
\begin{subfigure}{0.32\textwidth}
    \centering
    \includegraphics[width = \textwidth]{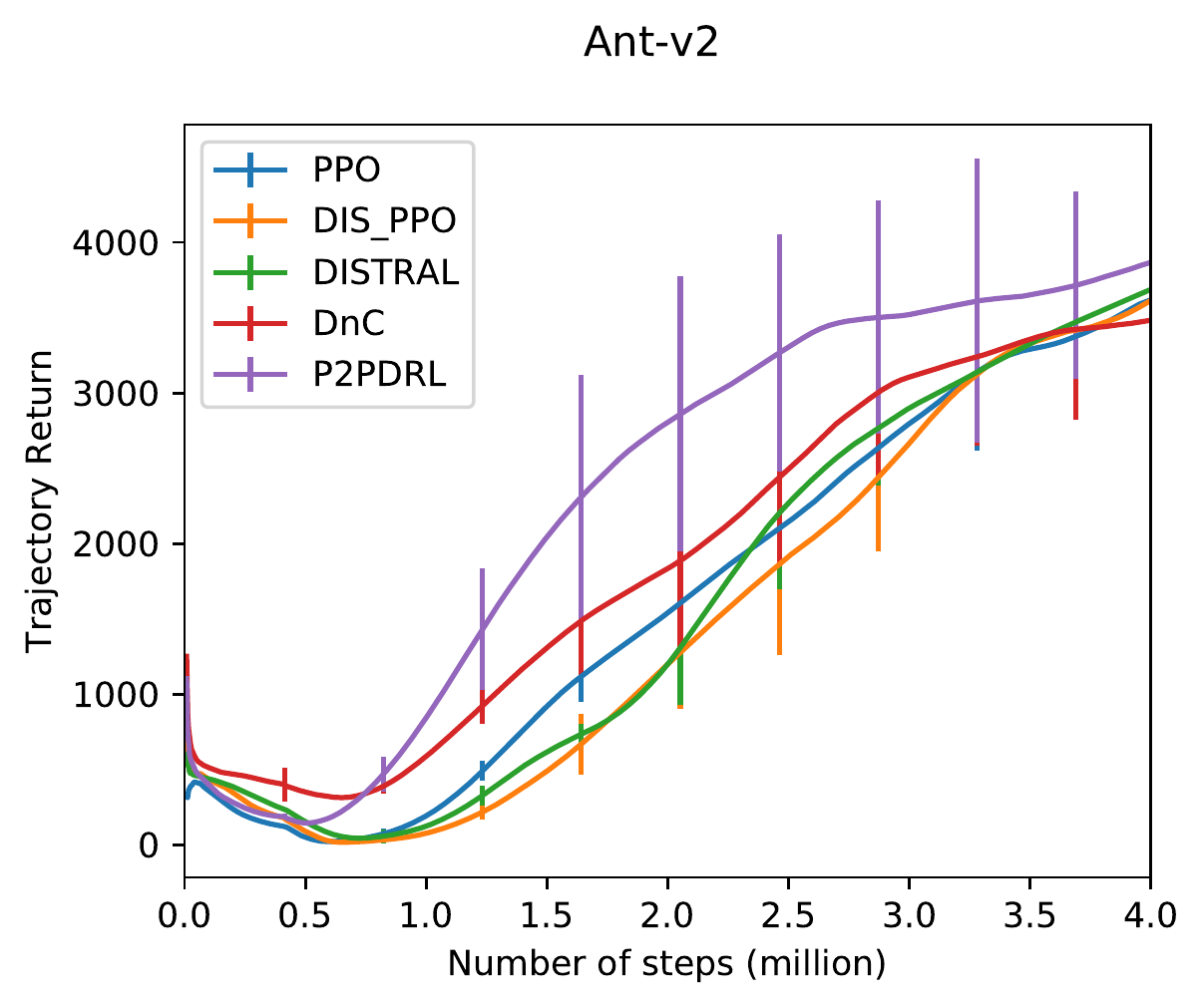}
    \caption{Training return on Ant}
\end{subfigure}
\begin{subfigure}{0.32\textwidth}
    \centering
    \includegraphics[width = \textwidth]{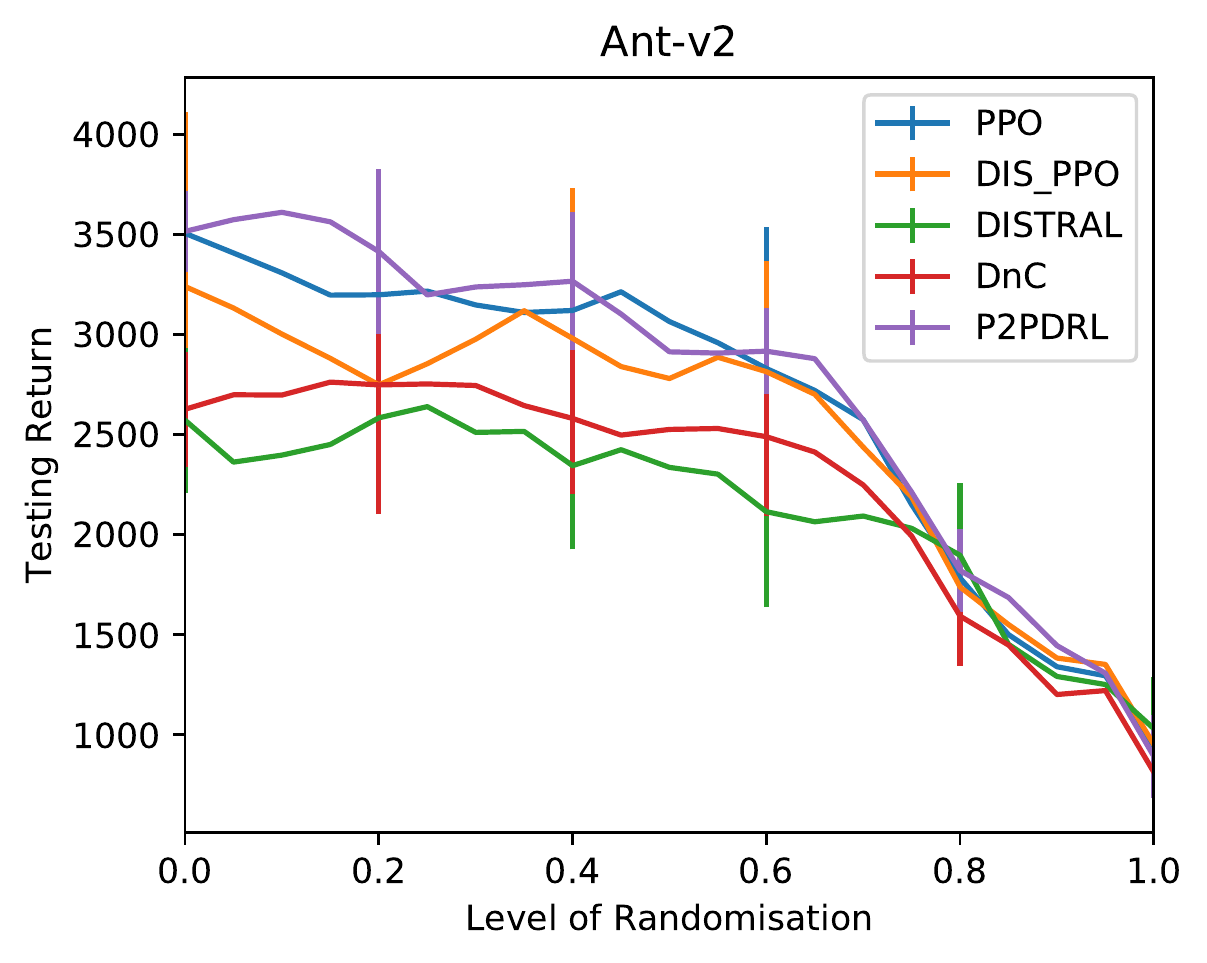}
    \caption{Testing return on Ant}
\end{subfigure}

\begin{subfigure}{0.32\textwidth}
    \centering
    \includegraphics[width = 0.75\textwidth]{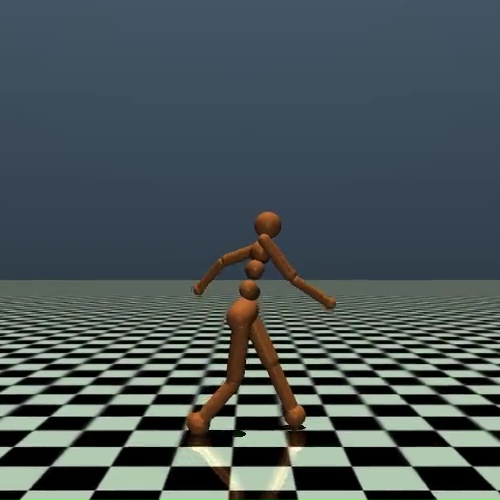}
    \caption{Humanoid task}
\end{subfigure}
\begin{subfigure}{0.32\textwidth}
    \centering
    \includegraphics[width = \textwidth]{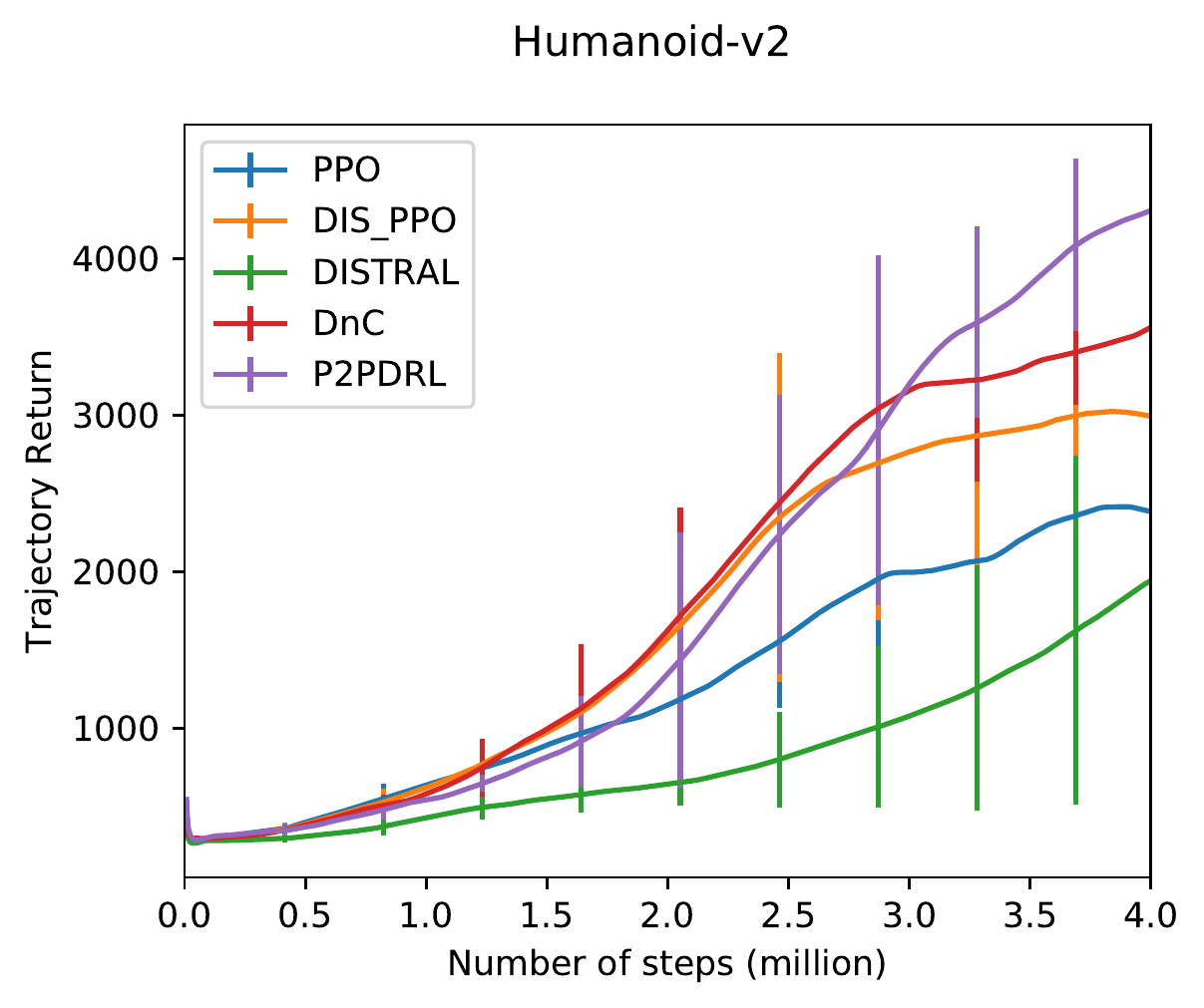}
    \caption{Training return on Humanoid }
\end{subfigure}
\begin{subfigure}{0.32\textwidth}
    \centering
    \includegraphics[width = \textwidth]{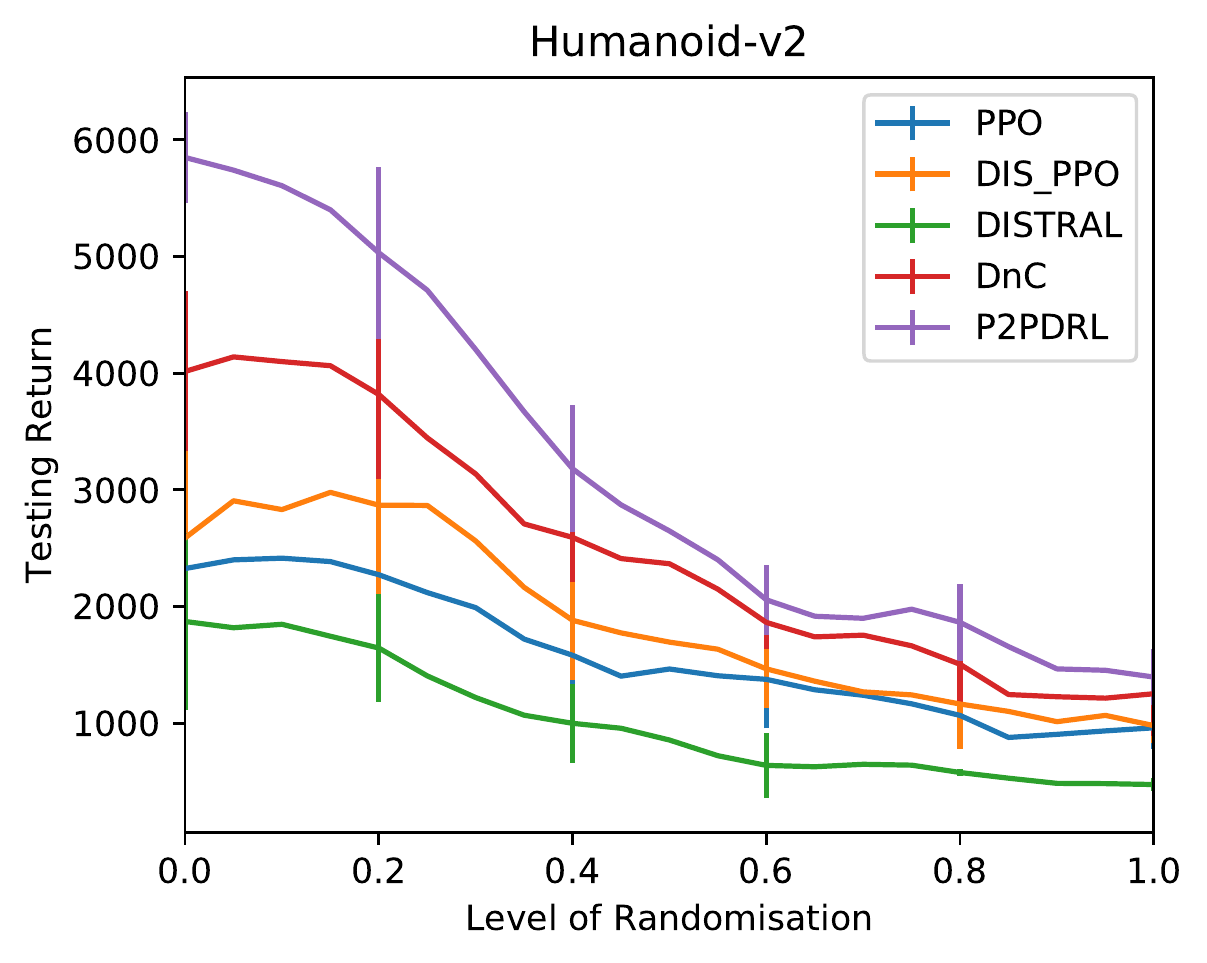}
    \caption{Testing return on Humanoid}
\end{subfigure}

\begin{subfigure}{0.32\textwidth}
    \centering
    \includegraphics[width = 0.75\textwidth]{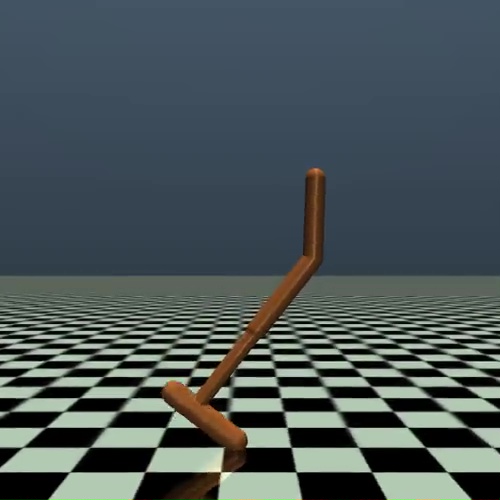}
    \caption{Hopper task}
\end{subfigure}
\begin{subfigure}{0.32\textwidth}
    \centering
    \includegraphics[width = \textwidth]{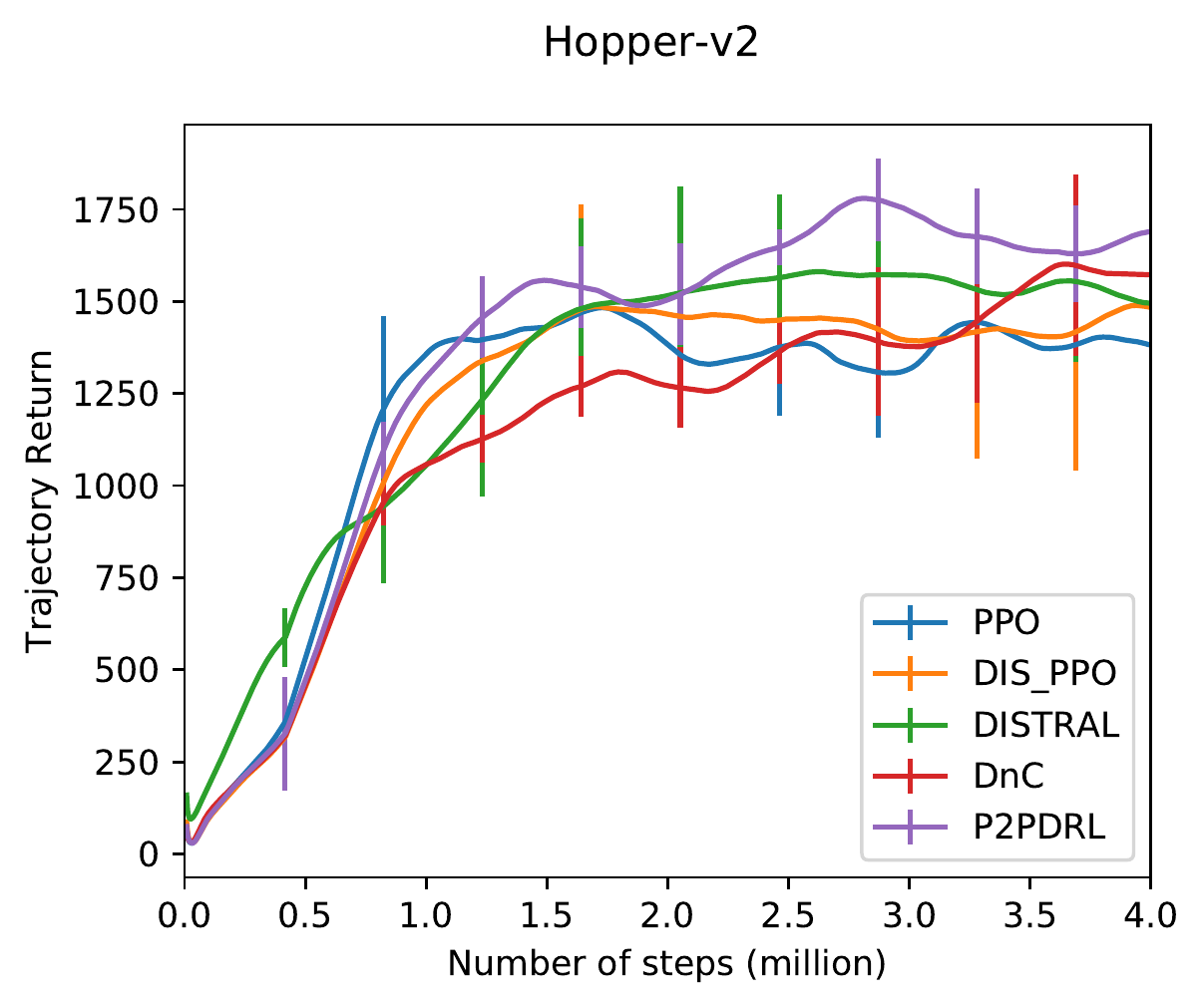}
    \caption{Training return on Hopper}
\end{subfigure}
\begin{subfigure}{0.32\textwidth}
    \centering
    \includegraphics[width = \textwidth]{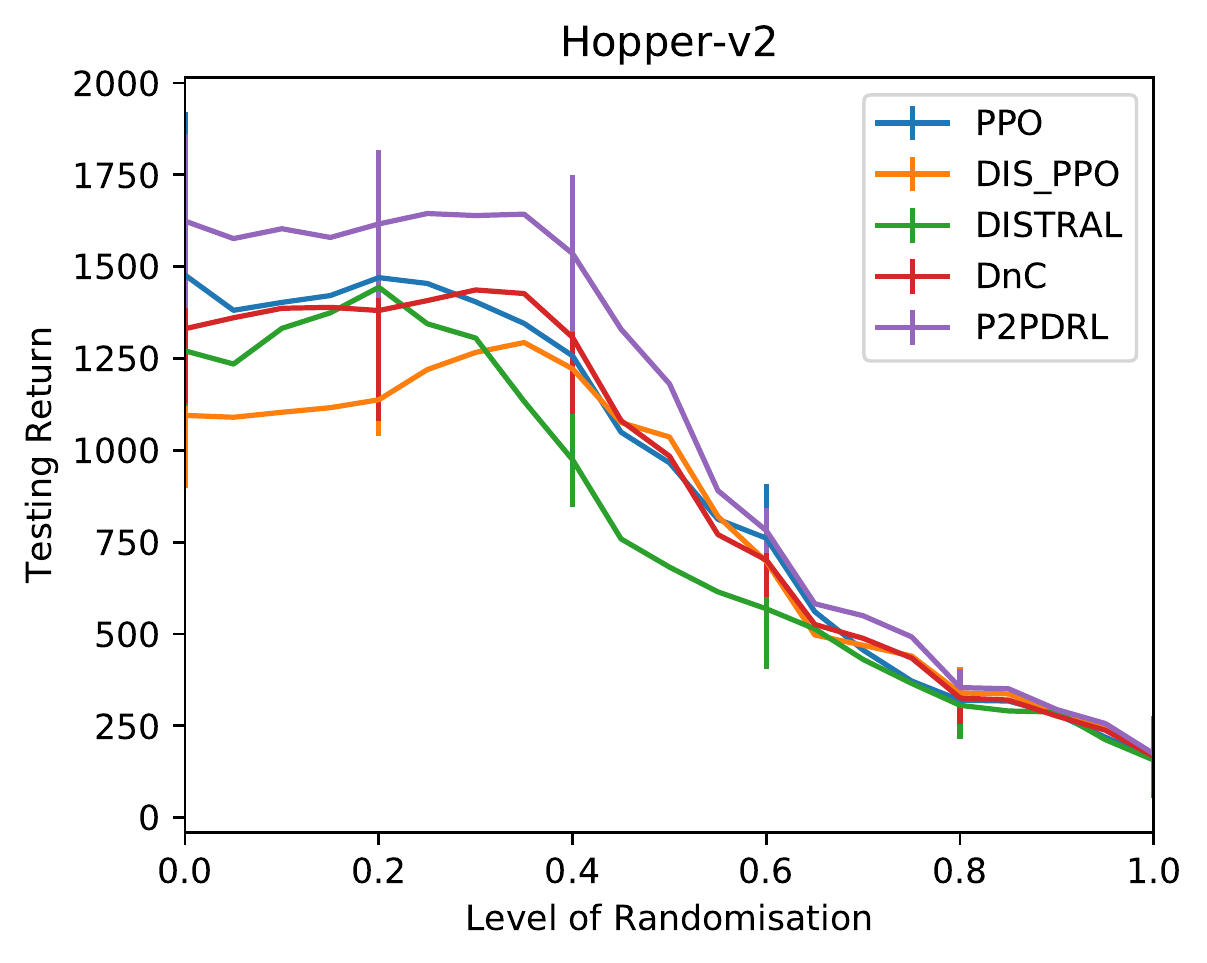}
    \caption{Testing return on Hopper}
\end{subfigure}

\begin{subfigure}{0.32\textwidth}
    \centering
    \includegraphics[width = 0.75\textwidth]{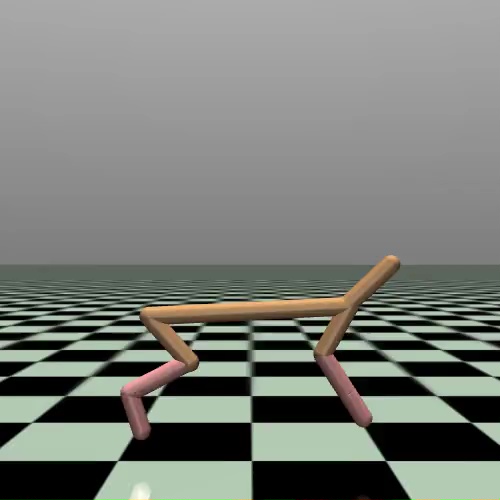}
    \caption{HalfCheetah task}
\end{subfigure}
\begin{subfigure}{0.32\textwidth}
    \centering
    \includegraphics[width = \textwidth]{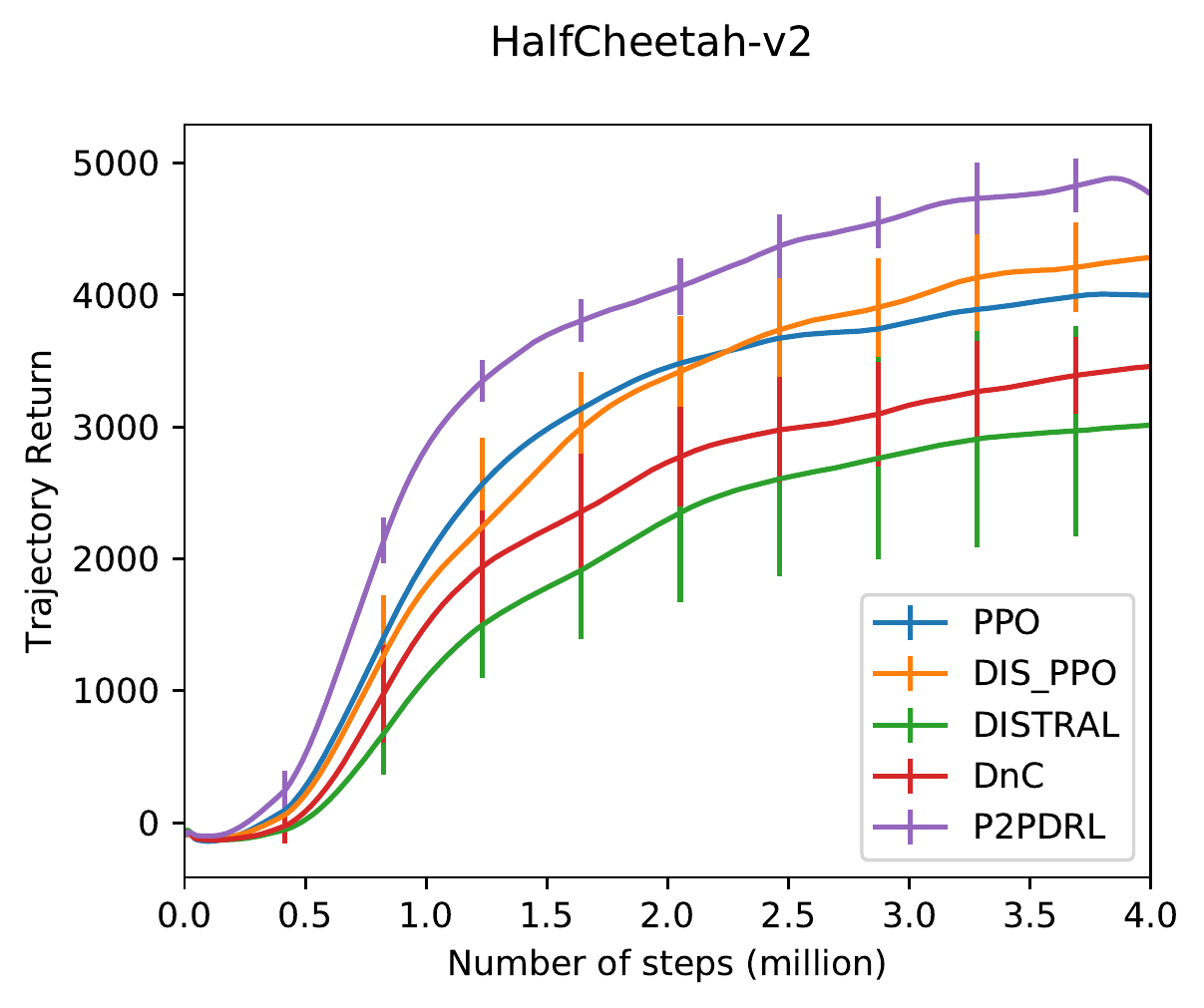}
    \caption{Training return on HalfCheetah}
\end{subfigure}
\begin{subfigure}{0.32\textwidth}
    \centering
    \includegraphics[width = \textwidth]{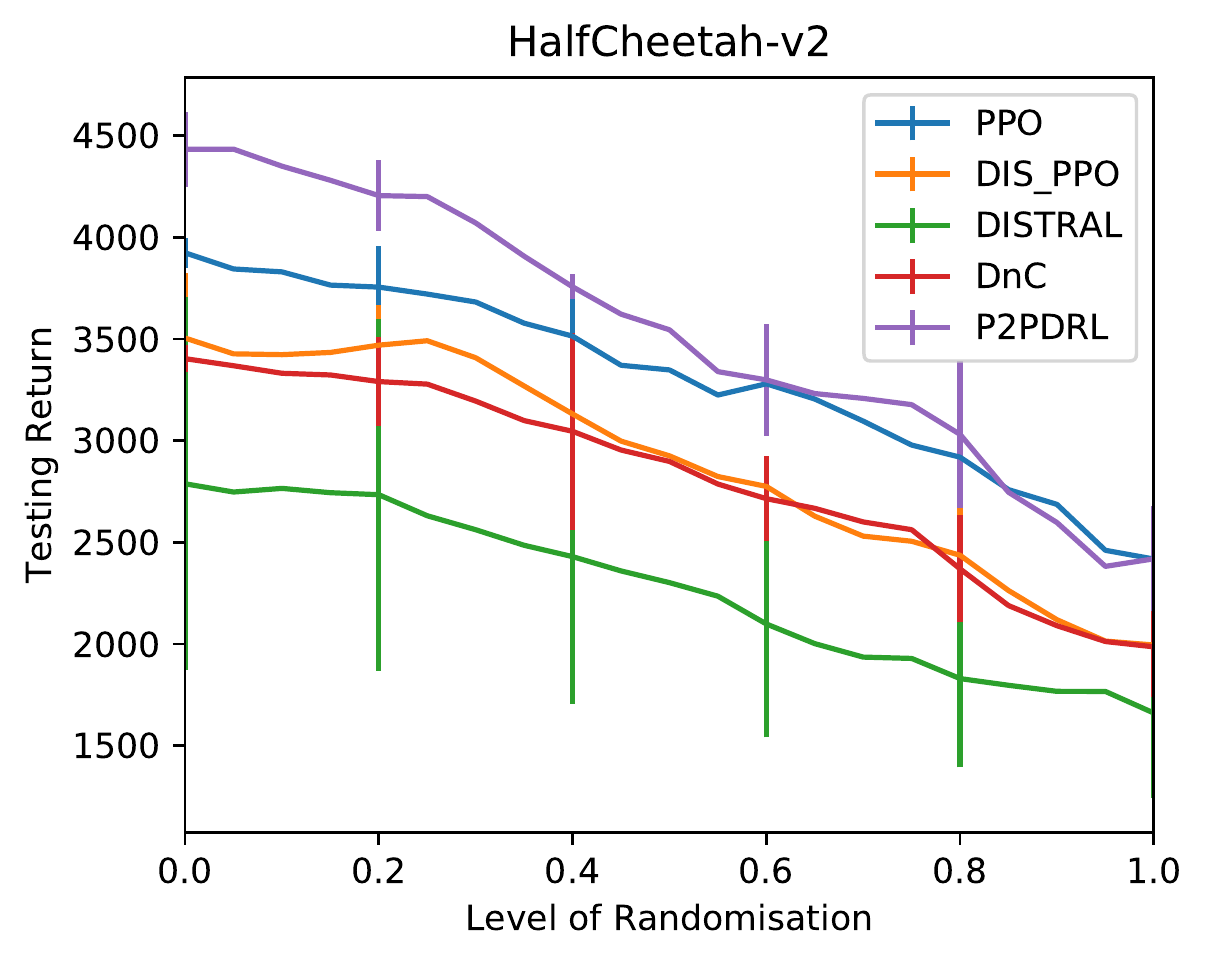}
    \caption{Testing return on HalfCheetah}
\end{subfigure}
\caption{Summary of training and generalisation performances across five different continuous control tasks. The leftmost column illustrates the task. The middle column compares the training return, represented by learning curves. The rightmost column shows the testing performance as a function of the testing randomised distribution diversity $\epsilon^{te}$. All experiments are trained with a predefined training randomisation distribution  $\epsilon^{tr}=0.2$. Results are averaged over 8  random seeds.}
\label{fig: table fig}
\end{figure}


%% file: supplementary.tex
\subsection*{Task Description}
For our experiments, we modify five continuous control tasks from OpenAI gym, 
including \textit{Walker2d}, \textit{Ant}, \textit{Humanoid}, \textit{Hopper} and \textit{HalfCheetah}. We use a scalar $\epsilon \in [0,1]$ to control the diversity of randomised domains. The pre-defined distributions are listed in Table.~\ref{tab: random dyna}.
\begin{table}[h]
    \vspace{-0.1cm}
    \caption{List of randomised distributions}
    \label{tab: random dyna}
    \centering
    \begin{tabular}{c c}
    \toprule 
        Wind condition & $w \sim \mathcal{U}(-5.0\epsilon, 5.0\epsilon)$  \\
        Gravity constant& $g \sim \mathcal{U}\big(g_0(1- 0.25\epsilon), g_0(1+0.25\epsilon)\big)$\\
        Friction coefficient & $f \sim  \mathcal{U}\big(f_0(1- 0.3\epsilon), f_0(1+0.3\epsilon)\big)$\\ 
        Robot mass & $m \sim \mathcal{U}\big(m_0(1- 0.5\epsilon), m_0(1+0.5\epsilon)\big)$ \\ 
        Initial position & $x \sim \mathcal{U}(x_0 - \epsilon, x_0+\epsilon)$  \\
    \bottomrule
    \vspace{-0.2cm}
    \end{tabular}
\end{table}
\vspace{-0.1cm}
\subsection*{Hyperparameters}
The detailed hyperparameters are listed below:
\begin{itemize}
    \item Network: 2 separate MLP networks for actor and critic, each with 2 hidden layer and 64 hidden units each layer, tanh nonlinearities. 
    \item Number of epochs each iteration: 10
    \item Minibatch size: 64
    \item Clipping parameter: 0.2
    \item Discount factor: 0.99
    \item GAE parameter: 0.95
    \item Optimiser: Adam optimiser
\end{itemize}
For PPO and Distributed PPO, we run a hyperparameter sweep on learning rate $\beta: [1\mathrm{e}-4,~3\mathrm{e}-4,~1\mathrm{e}-3,~3\mathrm{e}-3,~1\mathrm{e}-2]$. For Distral, DnC and P2PDRL, we run a hyperparameter sweep on learning rate $\beta: [1\mathrm{e}-4, 3\mathrm{e}-4, 1\mathrm{e}-3, 3\mathrm{e}-3, 1\mathrm{e}-2]$ and distillation loss coefficient $\alpha: [0.1, 0.3, 1.0, 3.0, 10.0]$. We report the best performing set of hyperparameters. 